\documentclass[sigconf]{acmart}
\AtBeginDocument{%
  }

\setcopyright{acmlicensed}
\copyrightyear{2018}
\acmYear{2018}
\acmDOI{XXXXXXX.XXXXXXX}
\acmConference[Conference acronym 'XX]{Make sure to enter the correct
  conference title from your rights confirmation email}{June 03--05,
  2018}{Woodstock, NY}
\acmISBN{978-1-4503-XXXX-X/2018/06}

\usepackage{multirow}
\usepackage{subfigure}
\usepackage{bm}
\usepackage[normalem]{ulem}
\useunder{\uline}{\ul}{}
\usepackage{algorithm}
\usepackage[noend]{algpseudocode}
\usepackage{enumitem}
\usepackage{graphicx}

\begin{document}

\title{UrbanDS: A Graph-Guided LLM Multi-Agent System for Data-Intensive Urban Tasks}


\author{Zhilun Zhou}
\affiliation{%
  \institution{Department of Electronic Engineering, BNRist, Tsinghua University}
  \city{Beijing}
  \country{China}}
\email{zzl22@mails.tsinghua.edu.cn}

\author{Jianghao Yu}
\affiliation{%
  \institution{Department of Electronic Engineering, BNRist, Tsinghua University}
  \city{Beijing}
  \country{China}}
\email{yujh2703@gmail.com}

\author{Yuming Lin}
\affiliation{%
  \institution{Department of Urban Planning, Tsinghua University}
  \city{Beijing}
  \country{China}}
\email{yujh2703@gmail.com}

\author{Yongjun Yang}
\affiliation{%
  \institution{Jiangmen Municipal Smart Social Governance Technology Innovation Center}
  \city{Jiangmen}
  \country{China}}
\email{muyigmail@gmail.com}

\author{Yongquan Sun}
\affiliation{%
  \institution{Tsinghua University}
  \city{Beijing}
  \country{China}}
\email{syq23@mails.tsinghua.edu.cn}

\author{Depeng Jin}
\affiliation{%
  \institution{Department of Electronic Engineering, BNRist, Tsinghua University}
  \city{Beijing}
  \country{China}
}
\email{jindp@tsinghua.edu.cn}

\author{Yong Li}
\correspondingauthor
\affiliation{%
  \institution{Department of Electronic Engineering, BNRist, Tsinghua University}
  \city{Beijing}
  \country{China}
}
\email{liyong07@tsinghua.edu.cn}


\begin{abstract}
Large language model (LLM) agents have been widely applied in various domains, with numerous studies exploring their applications in automating data science tasks. However, existing methods typically rely on a limited set of provided datasets, and they face challenges in data-intensive scenarios that require discovering and leveraging relevant information from large-scale and heterogeneous data repositories. Urban tasks are representative examples of such scenarios, as urban data are not only large-scale and multi-sourced, but also exhibit complex spatial, temporal, and semantic relationships across datasets. To address these challenges, we propose UrbanDS, a graph-guided LLM multi-agent system for data-intensive urban tasks. We first construct a unified dataset graph to organize reusable dataset skills and the relationships among datasets. Specifically, we develop a Data Profiling Agent that constructs a skill for each dataset, describing the content, schema, statistics, and usage. Moreover, a Relation Agent identifies relationships among datasets and integrates these relationships into the dataset graph. At runtime, a Planner Agent retrieves task-relevant datasets and relationships from the graph based on user queries and generates execution plans. Multiple Execution Agents then collaboratively perform code-based data processing and analysis, while their execution progress and intermediate results are shared through a common memory. Finally, a Report Agent synthesizes all the experimental logs into a report, which can be further refined based on user feedback. To systematically evaluate the capability of LLM agents in handling data-intensive urban scenarios, we further construct UrbanDS-Bench, an urban data science benchmark covering representative data analysis and modeling tasks. Experiments on both general and urban benchmarks demonstrate that UrbanDS consistently outperforms existing data science agents on data-intensive tasks. Furthermore, UrbanDS has been deployed on the urban operations platform of Dongxihu District, Wuhan, demonstrating its effectiveness in real-world urban applications. Our code and benchmark are released at https://github.com/zhou-zl18/UrbanDS.
\end{abstract}

\maketitle

\section{Introduction}

Large language models (LLMs) are rapidly evolving from passive assistants into autonomous agents that can reason, use tools, write and execute code, and coordinate with other agents. This development creates a promising way to automate data science workflows, which traditionally require substantial human effort in data understanding, preprocessing, analysis, modeling, and interpretation. Consequently, many existing studies have developed LLM agent systems for data science tasks ranging from data analysis to data modeling and machine learning~\cite{hollmann2023large,guo2024ds,hong2025data,trirat2025automl,zhang2025deepanalyze}. 

Existing data science agents usually rely on one or a few datasets provided directly with each task. However, in real-world applications, a data repository may contain a large number of heterogeneous datasets, most of which are irrelevant to a given task. Therefore, an agent must not only generate analysis code, but also discover relevant datasets from large-scale repositories before performing data processing and modeling. This challenge is particularly pronounced in urban scenarios. Urban data are typically multi-source and heterogeneous, covering domains such as mobility, transportation, population, land use, points of interest, economy, and public services. Moreover, these datasets often exhibit complex spatial, temporal, and semantic relationships, making it difficult for agents to identify relevant data sources and correctly integrate them for downstream analysis~\cite{zhang2026codabench}.

To address these limitations, in this work, we propose \textbf{UrbanDS}, a graph-guided LLM multi-agent system for data-intensive urban tasks. UrbanDS consists of two stages: dataset graph construction and task execution. In the first stage, we organize the datasets into a graph to model their attributes and relationships. Specifically, a Data Profiling Agent explores each dataset via Python code and creates a reusable \emph{dataset skill} that describes the dataset's content, schema, statistics, spatiotemporal coverage, and usage. Based on these, a Relation agent discovers relations between fields from different datasets via a semantic codebook. Moreover, we identify spatial and temporal relations from geographic coverage and time ranges between datasets. Finally, UrbanDS organizes the dataset skills and relations into a dataset graph, which helps agents quickly understand what each dataset contains, how to use it, and which other datasets may be relevant.
In the task execution stage, a Planner Agent retrieves relevant datasets from the dataset graph and creates a step-by-step analysis plan. Then, Execution Agents carry out the steps by writing and running code with a shared progress memory. After all steps are completed, a Reporter Agent generates the final report, which can be further revised based on user feedback.

To evaluate UrbanDS's ability in data-intensive urban tasks, we construct a benchmark for urban data science named \textbf{UrbanDS-Bench}. Specifically, we collect 94 datasets from ten major cities in China, covering geospatial, mobility, and socioeconomic data. Based on these, we construct 90 data analysis tasks and eight data modeling tasks, which comprehensively evaluate agents' capability of discovering and utilizing data from large-scale repositories.

We evaluate our method on UrbanDS-Bench and CoDA-Bench~\cite{zhang2026codabench}. Results show that UrbanDS surpasses existing data science agents as well as general coding agents by over 10\%, demonstrating its capability to handle data-intensive urban tasks. Moreover, UrbanDS has been deployed on the urban operations platform of Dongxihu District, Wuhan, where it supports practical analysis over real municipal data repositories. User studies further show that our system can significantly accelerate human data analysis processes.

Our main contributions are summarized as follows:
\begin{itemize}[leftmargin=*]
    \item We propose UrbanDS, a LLM multi-agent system for data-intensive urban tasks. We organize large-scale datasets into an easy-to-use graph, and further design a workflow that automatically discovers datasets, creates plans, executes code and generates reports.
    \item We construct UrbanDS-Bench to evaluate agents' ability to handle data-intensive urban tasks. It consists of 94 datasets from ten Chinese cities, 450 data analysis instances, and eight data modeling tasks.
    \item We conduct extensive experiments on two benchmarks, demonstrating the effectiveness of our method. Moreover, we deploy UrbanDS in the real-world to support urban data analysis.
\end{itemize}

\section{Related Work}
\label{sec:related work}

\subsection{LLM Agents for Data Science}
In recent years, there has been a growing interest in using LLM agents for data science tasks. DS-Agent uses case-based reasoning to adapt expert solutions and improve machine learning experiments through feedback~\cite{guo2024ds}. Data Interpreter represents a complex workflow as a hierarchical graph and progressively verifies and refines its execution~\cite{hong2025data}. AutoML-Agent coordinates specialized agents to cover the full AutoML pipeline from data retrieval to model deployment~\cite{trirat2025automl}, while LAMBDA employs programmer and inspector agents to make data analysis accessible through natural-language interaction~\cite{sun2026lambda}. More recently, DeepAnalyze trains an 8B agentic model to complete the process from raw data sources to analytical reports without relying on a fixed workflow~\cite{zhang2025deepanalyze}. These systems show the potential of LLMs to support data understanding, code generation, analysis, modeling, and reporting.

Progress in data science agents has also motivated benchmarks at different levels of complexity. DS-1000 and LLM4DS focus primarily on data science code generation~\cite{lai2023ds,nascimento2024llm4ds}. DSBench, DABstep, and DataSciBench further evaluate realistic analysis, modeling, multistep reasoning, and executable workflow completion~\cite{jing2025dsbench,egg2025dabstep,zhang2026datascibench}. However, these benchmarks generally provide agents with the datasets needed for each task, so they do not directly evaluate dataset discovery from a large repository. CoDA-Bench takes an important step toward this setting by placing code agents in noisy file systems that contain hundreds of datasets~\cite{zhang2026codabench}. Its results reveal that strong agents still struggle to combine data discovery with correct code execution. UrbanDS addresses this capability through reusable dataset skills, explicit dataset relations, and graph-guided task execution.

\begin{figure*}[ht]
    \centering
    \includegraphics[width=.99\linewidth]{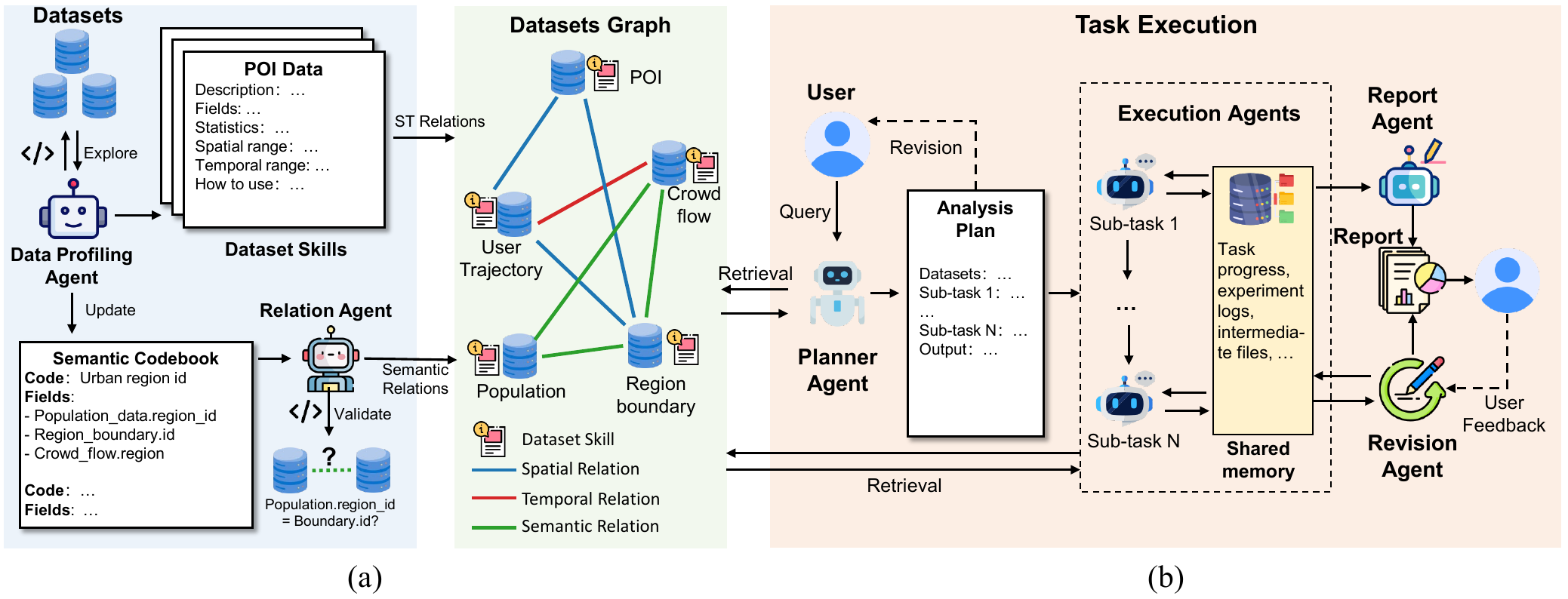}
    \vspace*{-10px}
    \caption{The overall framework of UrbanDS, including (a) dataset graph construction, and (b) task execution.}
    \vspace*{-10px}
    \label{fig:framework}
\end{figure*}

\subsection{Evaluation of LLM Agents on Urban Tasks}

Several benchmarks have evaluated LLMs as urban agents. CityBench integrates urban data with an interactive simulator and evaluates perception, understanding, and decision making across representative urban tasks and cities~\cite{feng2025citybench}. USTBench focuses on spatiotemporal reasoning and provides fine-grained evaluation of understanding, forecasting, planning, and reflection in an interactive city environment~\cite{lai2026ustbench}. These benchmarks demonstrate both the promise of LLMs for urban applications and their limitations in professional numerical reasoning, long-horizon planning, and adaptation.

Our benchmark draws inspiration from their design of representative urban tasks, but evaluates a different capability. In CityBench and USTBench, the task environment and required observations are prepared for the agent, and the main goal is to assess reasoning or decision-making within that environment. In UrbanDS-Bench, the agent receives a question and access to a large data pool, without being told which datasets are required. It must discover the relevant datasets, understand their content, determine how to use them together, and execute the analysis. UrbanDS-Bench therefore evaluates urban data science in a data-intensive repository rather than the inherent urban knowledge or reasoning ability of an LLM alone.

\section{Preliminaries}
\label{sec:preliminaries}

In this section, we introduce the basic concepts and formally define the task studied in this work.

\paragraph{Dataset Repository.}
Let $\mathcal{D}=\{D_1,D_2,\ldots,D_n\}$ denote a dataset repository containing $n$ datasets. Each dataset $D_i$ can be a single file or a directory containing a collection of related files. Urban datasets commonly use formats such as CSV, GeoJSON, XLSX, and Shapefile, and may contain tabular, spatial, temporal, or spatiotemporal information. The datasets in $\mathcal{D}$ can differ in their formats, schemas, spatial coverage, temporal coverage, and levels of detail.

\begin{definition}[Data-intensive Question Answering Task]
Based on the dataset repository $\mathcal{D}$, a data-intensive question answering task is defined as a tuple $T_k=(Q_k,\mathcal{D}_k,A_k)$, where $Q_k$ is a question expressed in natural language, $\mathcal{D}_k\subseteq\mathcal{D}$ is the set of datasets required to answer the question, and $A_k$ is the expected answer. The goal of the task is to identify the relevant datasets $\mathcal{D}_k$ from the complete repository $\mathcal{D}$ and perform the necessary data analysis to produce the answer $A_k$, given only the question $Q_k$.
\end{definition}

This task is challenging for the following reasons. First, the number of datasets $n$ in the repository may be very large in real-world scenarios, making it difficult to correctly identify the relevant datasets. Second, there are complex relationships between datasets, and failing to consider these relationships can lead to selecting incorrect datasets or missing necessary ones.

\section{Methods}
\label{sec:methods}

\subsection{Framework Overview}

To address the aforementioned challenges, we design UrbanDS with two stages: dataset graph construction and task execution.
For large-scale dataset repositories, inspecting every dataset when doing each task is costly and may exceed the context window of LLMs. Therefore, UrbanDS instead explores each dataset once and stores the resulting knowledge for later use. Specifically, A Data Profiling Agent explores the dataset through code and summarizes its content and usage in a dataset skill. UrbanDS then identifies how datasets may be connected and organizes the skills and relations into a dataset graph.
In the task execution stage, UrbanDS leverages this graph to answer a user query. Specifically, a Planner Agent progressively inspects and selects promising datasets and decomposes the query into executable steps. Then an Execution Agent completes each step by writing and executing code. All steps communicate through a shared progress memory that records observations, findings, and generated files. Finally, a Report Agent summarizes the progress into a data analysis report, and a Revision Agent updates the report or its artifacts according to user feedback.
The overall framework of UrbanDS is presented in Figure~\ref{fig:framework}.

\subsection{Dataset Graph Construction}
To model the attributes of datasets and the complex relationships between them, we construct a dataset graph, which can be formulated as $\mathcal{G}=(\mathcal{V},\mathcal{E})$. Each node $v_i\in\mathcal{V}$ corresponds to a dataset $D_i$ and stores its dataset skill. The edge set $\mathcal{E}=\mathcal{E}_{\mathrm{spa}}\cup\mathcal{E}_{\mathrm{tem}}\cup\mathcal{E}_{\mathrm{sem}}$ contains spatial, temporal, and semantic field relations. The graph is built through the following two steps.

\subsubsection{Skill Extraction}
To avoid exploring every dataset for each question, we propose to explore each dataset only once and store its information into a reusable skill. Specifically, we construct a Data Profiling Agent to explore each dataset by writing Python code. The agent examines schemas, samples, value distributions, and spatial and temporal coverage. After obtaining a basic understanding of the dataset, it produces a dataset skill that summarizes the dataset content, field meanings, data scale, spatial and temporal ranges, loading method, and important usage notes. The skills give later agents practical knowledge of how to use the data without repeatedly opening the raw files.

\subsubsection{Relation Identification}

Urban datasets often contain spatial and temporal information. Therefore, we model three types of relations between datasets: spatial, temporal, and semantic relations. For spatial and temporal relations, we compare the corresponding coverage recorded in the dataset skills. We create a spatial relation when two datasets have overlapping spatial coverage and a temporal relation when their time ranges overlap.

Apart from spatial and temporal overlap, datasets may be connected through fields that refer to the same entities. We call these connections semantic relations. For example, \texttt{region\_id} in a population dataset may use the same region identifiers as \texttt{id} in a region boundary dataset. A semantic relation allows agents to recognize that the two datasets can be joined even though their field names are different.
It is promising to identify semantic relations using LLMs. However, it would be very costly to compare every pair of datasets. To address these, we propose to first store potential relations in a semantic codebook, and then verify each relation via an LLM agent.

As shown in Algorithm~\ref{alg:semantic-relation}, UrbanDS first constructs an incremental semantic codebook while profiling the datasets. Each code represents one type of entity identifier or reference, such as a region identifier. The Data Profiling Agent processes the datasets one by one. For each field that may connect to another dataset, it compares the field meaning and sample values with the existing codes. If the field matches an existing code, it is added to that code. Otherwise, the agent creates a new code to describe its meaning. As more datasets are profiled, the codebook grows to cover their fields without requiring a predefined ontology.

After the codebook has been constructed, UrbanDS processes each code with a Relation Agent. Fields assigned to the same code are possible connections, but their identifier systems may still be incompatible. The Relation Agent therefore checks their meanings and values, removes incorrect assignments, and determines how the validated fields can be connected. It then creates a semantic relation that records the involved datasets, their related fields, and a short connection description. 

\algrenewcommand\algorithmicrequire{\textbf{Input:}}
\algrenewcommand\algorithmicensure{\textbf{Output:}}
\begin{algorithm}[t]
\caption{Codebook and Semantic Relation Construction}
\label{alg:semantic-relation}
\begin{algorithmic}[1]
\Require Datasets $\{D_i\}$ and dataset skills $\{S_i\}$
\Ensure Codebook $\mathcal{C}$ and semantic relation set $\mathcal{E}_{\mathrm{sem}}$
\State $\mathcal{C}\gets\emptyset$
\State $\mathcal{E}_{\mathrm{sem}}\gets\emptyset$
\For{each dataset $D_i$}
    \For{each field $f$ in $D_i$}
        \State $c\gets\Call{ProfilingAgent}{D_i,f,S_i,\mathcal{C}}$
        \Statex \hspace{\algorithmicindent}\Comment{Match an existing code, create a new code, or return \textsc{None}}
        \If{$c\neq\textsc{None}$}
            \State $\mathcal{C}[c]\gets\mathcal{C}[c]\cup\{(D_i,f)\}$
        \EndIf
    \EndFor
\EndFor
\For{each code $c$ in $\mathcal{C}$}
    \State $\mathcal{E}_c\gets\Call{RelationAgent}{c,\mathcal{C}[c],\{S_i\}}$
    \State $\mathcal{E}_{\mathrm{sem}}\gets\mathcal{E}_{\mathrm{sem}}\cup\mathcal{E}_c$
\EndFor
\State \Return $\mathcal{C},\mathcal{E}_{\mathrm{sem}}$
\end{algorithmic}
\end{algorithm}

\subsection{Task Execution}
Given a user query, UrbanDS first retrieves the required datasets and builds an analysis plan, then executes the plan one step at a time. The two phases use the same dataset graph but require different levels of information. Planning needs a broad view of possible data sources, whereas execution needs detailed data access and actual runtime feedback.

\subsubsection{Dataset Retrieval and Planning}

Providing every dataset skill to the Planner Agent would fill its context with mostly irrelevant information, while providing only dataset names would not support reliable selection. UrbanDS therefore retrieves data progressively, moving from broad and compact information to detailed and focused information. The planner first sees the names and short descriptions of all datasets and selects promising candidates based on the user query. It then reads the complete skills of these candidates to understand their contents and usage.

When the planner inspects a dataset, it also sees compact summaries of its neighboring datasets in the graph. These relations may reveal useful data whose descriptions alone do not appear directly relevant. The planner can inspect the full skill of a useful neighbor and continue following relations until it has enough information. A graph edge is treated as a possible connection rather than proof that a dataset must be used, so the planner still verifies each choice from its skill. It finally produces an analysis plan containing the selected datasets and an ordered set of executable steps.

\subsubsection{Sub-task Execution}

Based on the execution plan, we assign one Execution Agent to each planned step. Focusing an agent on one concrete step reduces the complexity of code generation, while shared results preserve dependencies across the full analysis. 
Specifically, each Execution Agent writes and runs code, observes the actual output, and revises its approach when errors or unexpected values occur. The resulting observations and findings are added to a shared progress memory, while generated files are also recorded for later steps. Subsequent agents can therefore reuse earlier results instead of repeating the same computation.
Moreover, if the datasets selected by the planner are insufficient during execution, the agent can also retrieve additional datasets from the dataset graph.
\subsubsection{Report Generation and Revision}

After all steps are completed, the Report Agent uses the plan, progress memory, and generated artifacts to produce a report grounded in the executed analysis. A Revision Agent can then update the report or its artifacts according to user feedback. This allows users to refine the results without restarting the complete task.

\begin{figure*}[t]
    \centering
    \includegraphics[width=.9\linewidth]{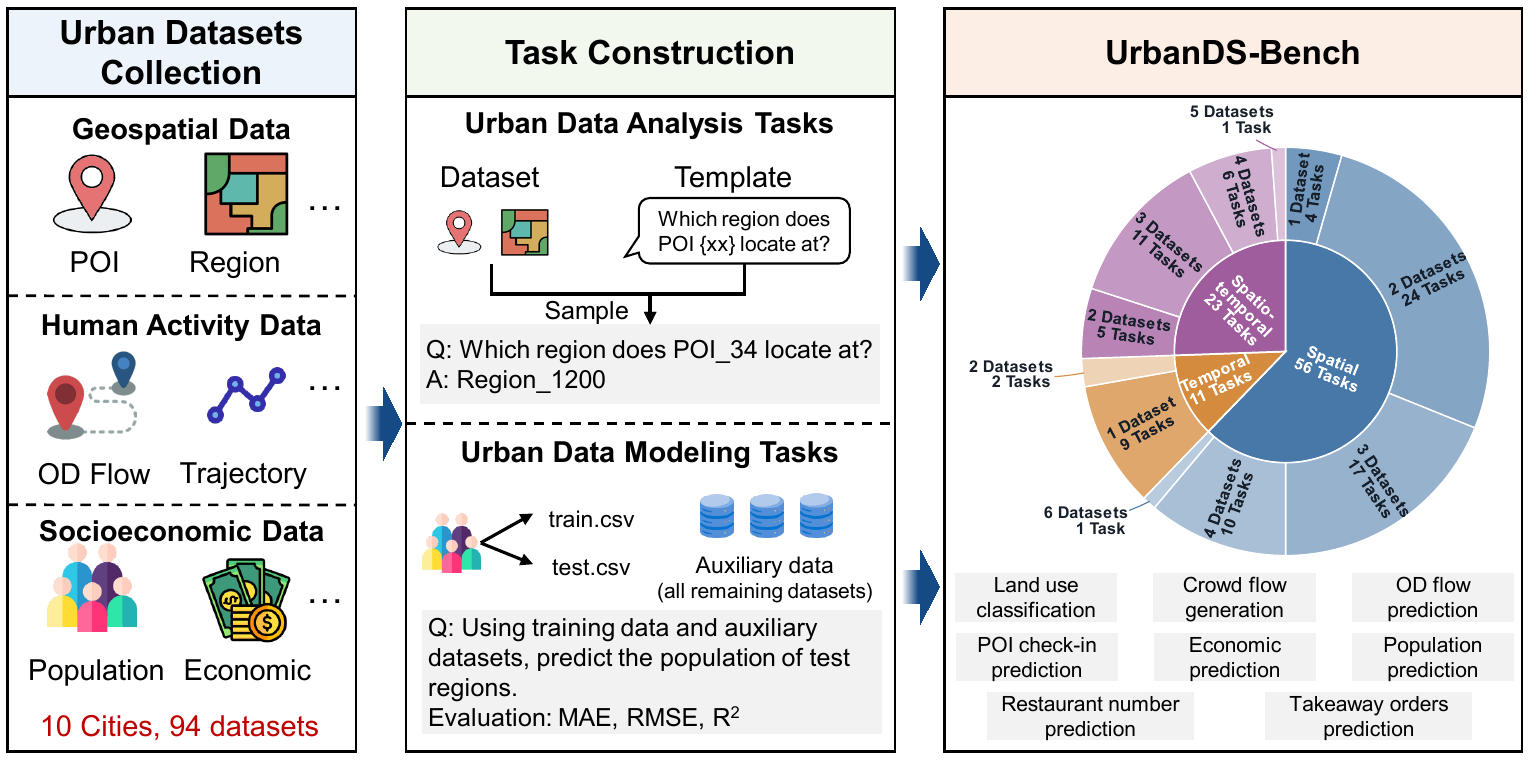}
    \vspace*{-10px}
    \caption{The construction and statistics of UrbanDS-Bench.}
    \vspace*{-10px}
    \label{fig:benchmark}
\end{figure*}

\section{Benchmark for Urban Data Science}

Existing data science benchmarks evaluate the ability of LLM agents to perform data analysis, visualization, and modeling~\cite{jing2025dsbench,egg2025dabstep,zhang2026datascibench,lai2023ds}. However, in these benchmarks, the datasets required by a task are directly provided, so an agent can begin by writing analysis code. As a result, these benchmarks do not evaluate whether an agent can discover a small set of relevant datasets from a large, heterogeneous repository and correctly connect them before analysis. This capability is essential in urban data science, where data are distributed across many sources and linked through nontrivial spatial, temporal, and semantic relations.

To address this limitation and systematically evaluate the capability of UrbanDS on data-intensive urban data science tasks, we construct a benchmark named UrbanDS-Bench. Specifically, we collect urban datasets from various sources for ten major Chinese cities and construct both data analysis and data modeling tasks. The overview of UrbanDS-Bench is shown in Figure~\ref{fig:benchmark}.

\subsection{Data Collection}
UrbanDS-Bench contains 94 datasets covering the 10 largest cities in China, including Beijing, Chongqing, Shanghai, Chengdu, Guangzhou, Shenzhen, Wuhan, Tianjin, Xi'an, and Zhengzhou. For every city, we collect eight core geospatial datasets from OpenStreetMap: administrative divisions, point-based points of interest (POIs), area-based points of interest (AOIs), buildings, roads, railways, waterways, and water bodies. These datasets provide a common basis for within-city and cross-city spatial analysis. We further collect human activity data and socioeconomic data to enrich the mobility and socioeconomic dimensions of the benchmark. They cover crowd inflow and outflow, OD flow, user trajectories, regional population, firm, restaurant, takeaway-order and price indicators, as well as POI category, brand, and commercial-area metadata. The resulting collection spans tabular, geospatial, and spatiotemporal data stored in CSV, GeoJSON, and JSON formats.

To create a genuinely data-intensive environment, we place all datasets in a single flat data pool, remove city names and most topic names from filenames, randomly shuffle the files, and assign anonymous identifiers. As a result, an agent generally cannot select a dataset from its filename alone; it must inspect file contents and schemas to infer the dataset semantics. The same pool also contains many plausible but irrelevant datasets for each task, requiring both dataset retrieval and valid cross-dataset integration.

\subsection{Data Analysis Tasks}
Following existing studies~\cite{lai2026ustbench,feng2025citybench}, we construct question--answer pairs that require executable analysis over the collected data. We organize them along two complementary dimensions: the spatiotemporal nature of the required reasoning and the number of datasets required to answer the question. The first dimension contains three categories. \emph{Spatial} tasks involve operations such as containment, intersection, proximity, buffering, and aggregation over geographic entities. \emph{Temporal} tasks require ordering, filtering, aggregation, or extrema detection over timestamps and time series. \emph{Spatiotemporal} tasks jointly reason about space and time, for example by mapping trajectories to regions or comparing time-varying flows among regions crossed by a transport feature. The second dimension ranges from one to six datasets and measures the breadth of dataset discovery and integration required by a task.

For each task type, we construct 5 questions by sampling from the data. For example, for the task of identifying the region of a POI ( Figure~\ref{fig:benchmark}), we randomly sample five POIs and compute the corresponding regions, resulting in five questions. 
Overall, the benchmark contains 90 task types and 450 questions. Among them, 56 task types (280 questions) are spatial, 11 types (55 questions) are temporal, and 23 types (115 questions) are spatiotemporal. Questions range from direct single-dataset aggregation to multi-stage analyses that combine as many as six independent files. Each question explicitly specifies the expected answer format, while its reference answer and required dataset set are generated from deterministic data operations. During evaluation, however, the agent is given the question and the full anonymized data pool rather than the identities of the required datasets.

\subsection{Data Modeling Tasks}
Beyond analytical QA, we construct eight Kaggle-style data modeling tasks, covering representative problems in urban computing. For each task, the agent receives a supervised training table, an unlabeled test table, and access to the remaining anonymized data pool. The agent needs to identify useful datasets, construct features, choose a validation strategy and model, and produce a submission in the specified CSV schema. The tasks are as follows.

\paragraph{Land use classification~\cite{xu2023spatial}} Given polygon boundaries and labels for training AOIs, the agent classifies held-out AOIs into nine land-use categories by exploiting their geometry and surrounding urban context. We report accuracy and macro-F1.
\paragraph{Crowd flow generation~\cite{zhou2023towards}} The agent generates a representative 24-hour weekday inflow and outflow profile for 362 spatially held-out regions after observing weekday flows in 648 other regions. We use MAE, RMSE, and SMAPE.
\paragraph{OD flow prediction~\cite{rong2024interdisciplinary}} We hide a stratified 20\% of zero and nonzero pairs in the $1{,}010\times1{,}010$ regional OD flow matrix and ask the agent to complete their flow values. We evaluate the predictions using the Common Part of Commuters (CPC) and MAE.
\paragraph{POI check-in prediction~\cite{zhang2026causalpoi}} The agent estimates check-in counts for held-out POIs using labels stratified by check-in magnitude and any non-target POI or urban-context data it discovers. We report MAE, RMSE, and $R^2$.
\paragraph{Economic prediction~\cite{zhou2023hierarchical}} The agent predicts firm counts for regions held out as complete spatial blocks, encouraging generalization beyond immediately neighboring labeled regions. We report MAE, RMSE, and $R^2$.
\paragraph{Population prediction~\cite{zhou2023hierarchical}} The agent estimates population for the same spatially held-out regions by constructing explanatory features from the available urban datasets. We report MAE, RMSE, and $R^2$.
\paragraph{Restaurant number prediction~\cite{zhou2023hierarchical}} The agent predicts the number of restaurants in spatially held-out regions from non-target urban context, evaluated by MAE, RMSE, and $R^2$.
\paragraph{Takeaway orders prediction~\cite{zhou2023hierarchical}} The agent predicts takeaway-order volume for spatially held-out regions; total order price is removed because it is a direct target-derived proxy. We report MAE, RMSE, and $R^2$.

Details about evaluation of data modeling tasks are presented in~\ref{app:modeling_metrics}.
For each modeling task, we split the corresponding target data into public training and test tables and treat the remaining files as candidate auxiliary datasets. We remove the original target file and any strongly target-derived source that could leak test labels. Unlike the data analysis tasks, a modeling task has no single ground-truth set of required auxiliary datasets: an agent may select and combine any leakage-safe sources that improve its model. This setting therefore evaluates whether an agent can make effective use of a large urban data repository under a more open-ended objective, in addition to training a predictive model.

\section{Experiments}
\label{sec:experiments}

\subsection{Experiment Settings}
\subsubsection{Benchmarks}

We evaluate UrbanDS on two benchmarks. \textbf{UrbanDS-Bench}, introduced in the previous section, includes 450 data analysis tasks and eight data modeling tasks. \textbf{CoDA-Bench}~\cite{zhang2026codabench} evaluates whether code agents can jointly perform data discovery and executable analysis in large file systems. We use its hard subset, which contains 119 challenging tasks from 15 Kaggle communities. Each task is placed in its complete community repository, requiring an agent to locate relevant resources among approximately 1,422 files on average. 

For evaluation metrics, we report answer accuracy for the data analysis tasks in UrbanDS-Bench and CoDA-Bench. The data modeling tasks use their corresponding predictive metrics introduced previously.     

\subsubsection{Baselines}
We compare UrbanDS with the following baselines:
\begin{itemize}[leftmargin=*]
    \item \textbf{DS-Agent}~\cite{guo2024ds}: A data science agent that uses case-based reasoning to adapt expert solutions from Kaggle and improves generated experiments through iterative feedback.
    \item \textbf{Data Interpreter}~\cite{hong2025data}: An LLM agent that represents complex data science workflows with a hierarchical graph and progressively verifies and refines each step during execution.
    \item \textbf{DeepAnalyze}~\cite{zhang2025deepanalyze}: An 8B agentic model trained with a curriculum of data science tasks to complete the full process from raw data sources to analytical reports.
    \item \textbf{AutoGen}~\cite{wu2023autogen}: A general multi-agent framework in which customizable agents communicate through natural language and code and use tools to complete a task.
    \item \textbf{Claude Code}\footnote{\url{https://claude.com/product/claude-code}}: A general-purpose coding agent from Anthropic that can inspect files, write code, and execute commands in a working environment.
\end{itemize}
These baselines cover state-of-the-art data science agents, a general multi-agent framework, and a commercial coding agent. This range allows us to compare UrbanDS with both systems designed for data science and agents designed for general coding tasks.

\subsubsection{Implementation Details}
DeepAnalyze is an open-source agentic model with 8B parameters, so we use its official model and implementation.
For all other baselines and our method, we use DeepSeek-V4-Pro~\cite{xu2026deepseek} with a temperature of 0 to ensure fair comparison.  All methods receive the same task instruction, dataset repository, execution environment, and required answer format.

\subsection{Overall Performance}
\begin{table*}[ht!]
    \centering
    \caption{Performance comparison with baselines on UrbanDS-Bench and CoDA-Bench. The best results are in bold, and the second-best results are underlined.}
    \resizebox{\textwidth}{!}{
\begin{tabular}{c|cccc|c|cc|ccc|ccc}
\hline
\multirow{2}{*}{\textbf{Method}} & \multicolumn{4}{c|}{\textbf{UrbanDS-Bench}}                                       & \textbf{CoDA-Bench} & \multicolumn{2}{c|}{\textbf{Land Use Classification}} & \multicolumn{3}{c|}{\textbf{Crowd Flow Generation}} & \multicolumn{3}{c}{\textbf{Check-in Prediction}} \\
                                 & \textbf{Spatial/\%} & \textbf{Temporal/\%} & \textbf{ST/\%} & \textbf{Overall/\%} & \textbf{Overall/\%} & \textbf{Acc}              & \textbf{F1}               & \textbf{MAE}    & \textbf{RMSE}   & \textbf{SMAPE}  & \textbf{MAE}   & \textbf{RMSE}  & \textbf{R$^2$} \\ \hline
\textbf{DeepAnalyze}             & 0.7                 & 0.0                  & 0.0            & 0.4                 & 0.8                 & 0.371                     & 0.319                     & 7.486           & 11.065          & 1.249           & 2.824          & 3.458          & -1.063         \\
\textbf{DS-Agent}                & 3.9                 & 18.2                 & 0.0            & 4.7                 & 2.5                 & 0.399                     & 0.332                     & 6.705           & 10.911          & 1.116           & 3.841          & 4.319          & -2.219         \\
\textbf{Data Interpreter}        & 4.6                 & 3.6                  & 0.0            & 3.3                 & 5.0                 & 0.381                     & 0.305                     & 2.000           & 5.762           & 0.651           & 3.724          & 4.199          & -2.043         \\
\textbf{AutoGen}                 & 40.4                & {\ul 76.4}           & 43.5           & 45.6                & 31.1                & {\ul 0.502}               & {\ul 0.436}               & 2.832           & 5.762           & 0.841           & 2.404          & {\ul 2.792}    & {\ul -0.345}   \\
\textbf{Claude Code}             & {\ul 62.5}          & 56.4                 & {\ul 67.0}     & {\ul 62.9}          & {\ul 42.0}          & 0.444                     & 0.386                     & {\ul 1.821}     & {\ul 5.681}     & {\ul 0.656}     & {\ul 2.111}    & 3.209          & -0.777         \\ \hline
\textbf{UrbanDS}                 & \textbf{65.7}       & \textbf{83.6}        & \textbf{73.9}  & \textbf{70.0}       & \textbf{46.2}       & \textbf{0.550}            & \textbf{0.483}            & \textbf{1.795}  & \textbf{5.622}  & \textbf{0.617}  & \textbf{0.942} & \textbf{1.763} & \textbf{0.464} \\ \hline
\end{tabular}
    }
    \label{tbl:main1}
\end{table*}

\begin{table*}[ht!]
    \centering
    \caption{Performance comparison with baselines on the remaining data modeling tasks of UrbanDS-Bench. The best results are in bold, and the second-best results are underlined.}
    \resizebox{\textwidth}{!}{
\begin{tabular}{c|cc|ccc|ccc|ccc|ccc}
\hline
\multirow{2}{*}{\textbf{Method}} & \multicolumn{2}{c|}{\textbf{OD Flow Prediction}} & \multicolumn{3}{c|}{\textbf{Economic Prediction}} & \multicolumn{3}{c|}{\textbf{Population Prediction}} & \multicolumn{3}{c|}{\textbf{Restaurant Prediction}} & \multicolumn{3}{c}{\textbf{Orders Prediction}}   \\
                                 & \textbf{CPC}            & \textbf{MAE}           & \textbf{MAE}    & \textbf{RMSE}  & \textbf{R$^2$} & \textbf{MAE}    & \textbf{RMSE}   & \textbf{R$^2$}  & \textbf{MAE}    & \textbf{RMSE}   & \textbf{R$^2$}  & \textbf{MAE}   & \textbf{RMSE}  & \textbf{R$^2$} \\ \hline
\textbf{DeepAnalyze}             & 0.147                   & 8.735                  & 2.403           & 2.951          & -0.818         & 0.836           & 1.148           & 0.495           & 0.793           & 1.032           & 0.716           & 3.608          & 5.171          & -0.725         \\
\textbf{DS-Agent}                & {\ul 0.645}             & {\ul 3.029}            & 1.273           & 1.677          & 0.413          & 1.113           & 1.511           & 0.124           & 0.953           & 1.260           & 0.576           & 1.962          & 3.068          & 0.393          \\
\textbf{Data Interpreter}        & 0.315                   & 6.933                  & 2.015           & 2.488          & -0.292         & 0.984           & 1.262           & 0.390           & 2.109           & 2.654           & -0.879          & 2.162          & 3.228          & 0.327          \\
\textbf{AutoGen}                 & 0.297                   & 5.192                  & 1.316           & 1.702          & 0.395          & {\ul 0.773}     & {\ul 1.075}     & {\ul 0.557}     & 0.841           & 1.152           & 0.646           & 2.115          & 3.256          & 0.316          \\
\textbf{Claude Code}             & 0.472                   & 4.116                  & {\ul 1.065}     & {\ul 1.314}    & {\ul 0.640}    & 0.792           & 1.094           & 0.541           & {\ul 0.656}     & {\ul 0.871}     & {\ul 0.797}     & {\ul 1.710}    & {\ul 2.484}    & {\ul 0.602}    \\ \hline
\textbf{UrbanDS}                 & \textbf{0.730}          & \textbf{2.519}         & \textbf{0.831}  & \textbf{1.103} & \textbf{0.746} & \textbf{0.681}  & \textbf{0.930}  & \textbf{0.669}  & \textbf{0.584}  & \textbf{0.805}  & \textbf{0.827}  & \textbf{1.629} & \textbf{2.361} & \textbf{0.640} \\ \hline
\end{tabular}
    }
    \label{tbl:main2}
\end{table*}

\begin{figure}[h]
    \centering
    \vspace{-10px}
    \includegraphics[width=.8\linewidth]{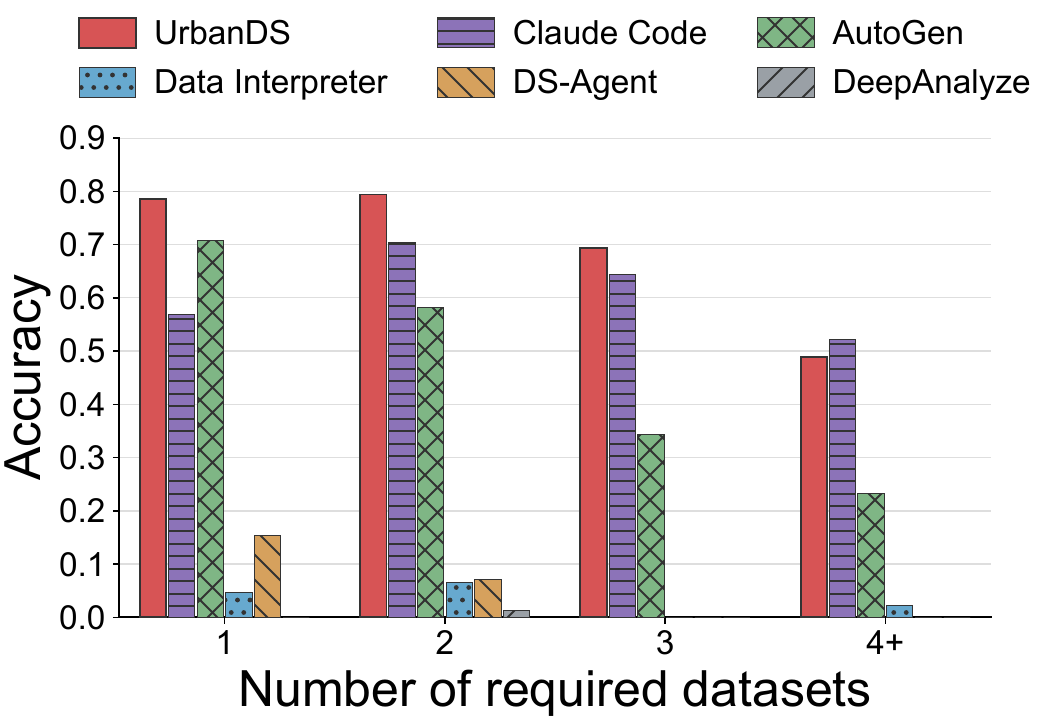}
    \vspace{-10px}
    \caption{Accuracy under different dataset numbers in UrbanDS-Bench.}
    \vspace{-10px}
    \label{fig:acc_by_dataset_num}
\end{figure}

Tables~\ref{tbl:main1} and~\ref{tbl:main2} report the overall results, from which we have the following findings.

First, UrbanDS achieves the best accuracy in every reasoning category of UrbanDS-Bench, reaching 65.7\% on spatial tasks, 83.6\% on temporal tasks, and 73.9\% on spatiotemporal tasks. Its overall accuracy is 70.0\%, compared with 62.9\% for the strongest baseline, Claude Code. This corresponds to a relative improvement of 11.2\%. UrbanDS also achieves 46.2\% accuracy on CoDA-Bench, improving upon Claude Code by 10.0\%. The consistent gains on both the urban benchmark and the general data discovery benchmark show that UrbanDS is effective beyond a single data domain.
UrbanDS also performs strongly on the eight data modeling tasks. It obtains the best result on all metrics. The improvement is especially clear on tasks that benefit from auxiliary data. For example, UrbanDS achieves an $R^2$ of 0.464 for POI check-in prediction, while all baselines produce negative $R^2$ values. These results indicate that UrbanDS can discover and use relevant datasets to build more accurate models.

Among the baselines, the data science agents DS-Agent and Data Interpreter perform poorly on UrbanDS-Bench and CoDA-Bench. This is probably because these agents are mainly designed to plan an analysis after the input data have already been specified. In our setting, however, an agent must first explore a large repository, identify the required datasets, and adjust its plan as new dataset information becomes available. General coding agents, especially Claude Code and AutoGen, handle this exploration more effectively, but still lack an explicit representation of dataset content and connections. UrbanDS addresses this difficulty with dataset skills, the dataset graph, and progressive retrieval by the Planner Agent. Additionally, DeepAnalyze performs worst on the two analysis benchmarks, suggesting that agentic training alone does not fully compensate for the limited capability of its 8B backbone on tasks that require extensive data discovery and reasoning.

Figure~\ref{fig:acc_by_dataset_num} further compares accuracy by the number of datasets required for each UrbanDS-Bench task. UrbanDS performs best on tasks involving one, two, or three datasets, with accuracies of 78.5\%, 79.4\%, and 69.3\%, respectively. When tasks require four or more datasets, the accuracy of every method is below 53\%. Performance generally becomes lower as more datasets must be jointly identified and used, particularly from three datasets onward. This result confirms that finding the correct data and combining information across datasets are central challenges in data-intensive analysis. Although UrbanDS substantially reduces this difficulty, tasks involving many datasets remain challenging.

\subsection{Ablation Study}
\begin{figure}[h]
    \centering
    \vspace{-10px}
    \includegraphics[width=.9\linewidth]{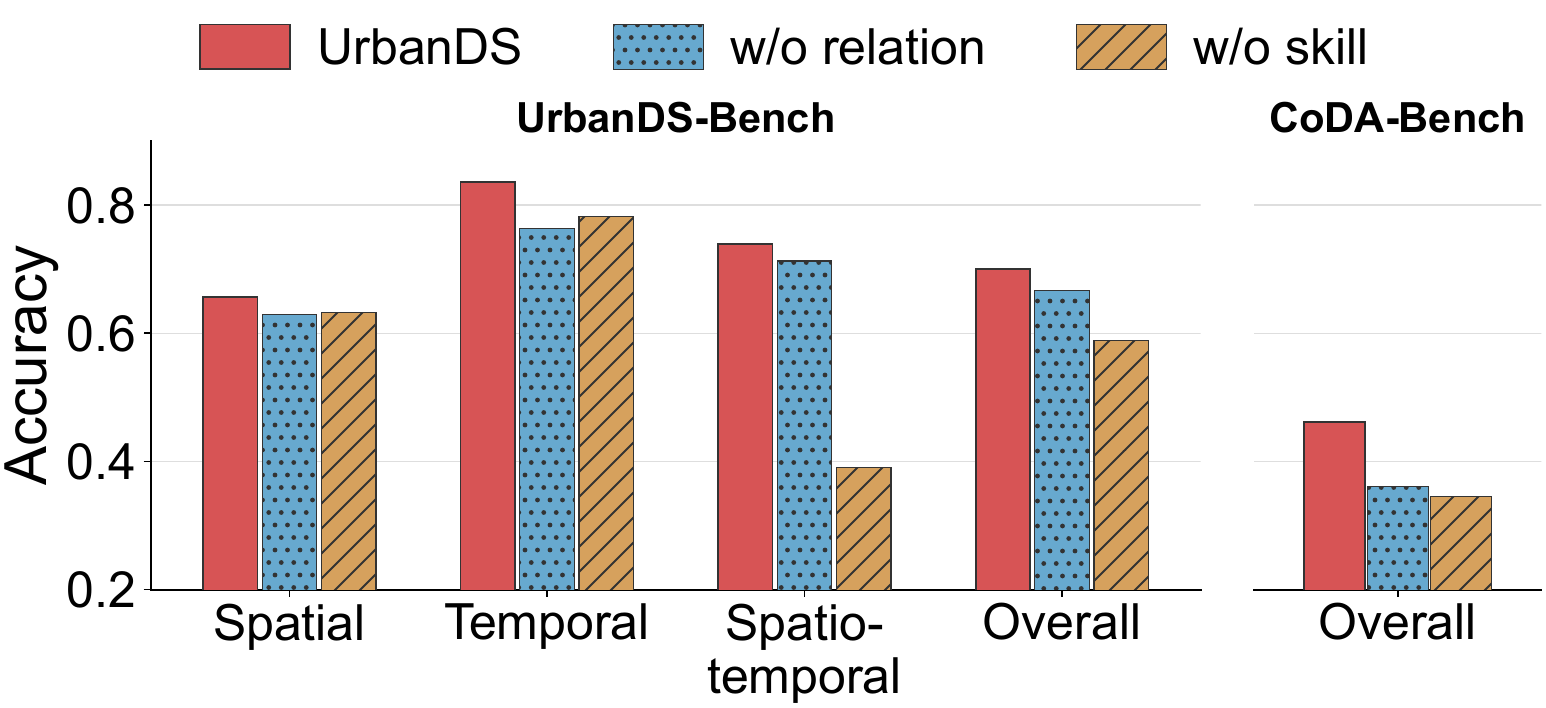}
    \vspace{-10px}
    \caption{Ablation study results.}
    \vspace{-10px}
    \label{fig:ablation}
\end{figure}

We evaluate two variants of UrbanDS to understand the contribution of dataset relations and dataset skills. The first variant removes the relations between datasets, so the Planner Agent cannot use the dataset graph to locate connected data. The second removes dataset skills and exposes the agent to the original files without the structured descriptions produced during data preparation.

As shown in Figure~\ref{fig:ablation}, both variants perform worse on UrbanDS-Bench and CoDA-Bench. Removing dataset relations reduces the overall accuracy from 70.0\% to 66.7\% on UrbanDS-Bench and from 46.2\% to 36.1\% on CoDA-Bench. The relative decrease is 13.3\% on average across the two benchmarks. This result shows that explicit relations help the agent move from an initially relevant dataset to other datasets needed by the task. The effect is particularly clear on CoDA-Bench, where useful resources are distributed across a large file repository.

Removing dataset skills causes a larger decrease. Accuracy falls to 58.9\% on UrbanDS-Bench and 34.5\% on CoDA-Bench, giving an average relative decrease of 20.6\%. The largest change within UrbanDS-Bench occurs on spatiotemporal tasks, where accuracy drops from 73.9\% to 39.1\%. These tasks require the agent to understand both the content and usage of heterogeneous datasets before combining them. The results therefore show that dataset relations mainly support data discovery, while dataset skills provide the detailed knowledge required to inspect, select, and correctly use the discovered data. Both components are important for reliable analysis over large dataset repositories.

\subsection{Deployment and User Study}
\subsubsection{Real-world Deployment}
\begin{figure}[h]
    \centering
    \includegraphics[width=.99\linewidth]{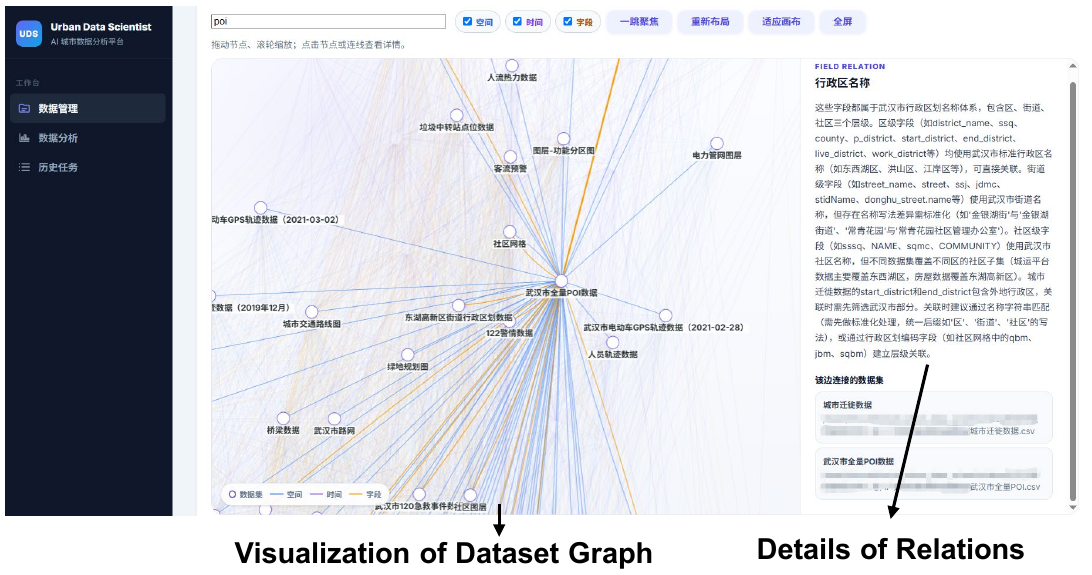}
    \caption{Interface of the deployed system.}
    \label{fig:deployment}
\end{figure}

To evaluate UrbanDS beyond controlled benchmarks, we deployed it on the urban operations platform of Dongxihu District, Wuhan. The deployed repository currently contains 103 dataset skills constructed directly from real urban and operational data. These datasets cover urban governance events, transportation and mobility, emergency response, population and communities, industrial economy, public facilities, municipal networks, environment, and land-use planning. They include heterogeneous tables, geographic layers, time series, trajectories, and text stored in formats such as CSV, Excel, GeoJSON, JSON, and SQL. This setting reflects a practical difficulty of urban operations: relevant data are distributed across many domains and formats, making manual discovery and integration costly.

The profiling and relation construction stages convert this data into reusable skills and a dataset graph. The current graph contains more than 2,200 spatial, temporal, and semantic relations, allowing the Planner Agent to discover useful data connections when answering natural-language requests. Figure~\ref{fig:deployment} shows an interface of the deployed platform that visualizes the dataset graph, where different colors represent different relation types. Users can conveniently search for a dataset, view its skill to understand its content and usage, and quickly locate related datasets through the graph. The deployment demonstrates that UrbanDS can organize a real urban data repository and make its data knowledge more accessible for practical analysis.

\subsubsection{User Study}

In addition to supporting urban operations, the deployed platform allows users to register their own datasets and use UrbanDS for their analysis tasks. We invited 12 users to use the platform for approximately one month and collected questionnaires and usage logs. The participants were mainly graduate researchers and urban professionals from fields including urban planning, transportation, economy, public services, environment, and population. They were familiar with urban analysis but had different levels of programming experience. By analyzing the questionnaires and usage logs, we have the following findings.

First, UrbanDS substantially reduced analysis time. According to participants' estimates, manually completing a data analysis task typically takes around 4.08 hours on average, whereas using UrbanDS reduces the required time to only 0.73 hours, corresponding to an estimated speedup of 5.6 times. 
Second, the tasks performed by users typically involved two to six datasets. Nine users usually worked with two or three datasets, while three users used four to six, which is consistent with the setting of up to six datasets in our UrbanDS-Bench. 
Third, users generally found UrbanDS useful, with eight participants rating its task completion and helpfulness at 4 or 5 out of 5. The logs also show that users inspected and revised the generated results, indicating that UrbanDS is most useful as an analysis assistant whose outputs remain open to human verification. 

\section{Conclusion}
\label{sec:conclusion}

In this paper, we presented UrbanDS, a multi-agent system for data-intensive urban data science tasks. UrbanDS first creates a reusable skill for each dataset and identifies spatial, temporal, and semantic relations between datasets, resulting in a dataset graph. Based on this graph, the Planner Agent progressively finds relevant datasets and constructs an analysis plan. Execution Agents then complete the plan through code execution and runtime feedback, while shared progress records connect the results of different steps. The system finally generates a report and supports further revision based on user feedback. We also construct UrbanDS-Bench to evaluate agents' ability to handle data-intensive urban tasks. Experiments show that UrbanDS outperforms existing data science agents and strong general coding agents. The real-world deployment and user study further demonstrate its practical value for organizing urban data and supporting users in complex analysis tasks.

Like existing data science agents, UrbanDS currently focuses on completing analysis and modeling tasks specified by users. It remains limited in more open-ended settings, such as exploring a repository without a predefined question, identifying unexpected patterns, and proposing new research questions from data~\cite{xia2025ai}. Future work will extend UrbanDS from executing user requests toward actively discovering and validating useful insights, with the broader goal of building a more capable urban data scientist.

\bibliographystyle{ACM-Reference-Format}
\bibliography{reference}

\appendix
\newpage
\section{Evaluation Metrics for Data Modeling Tasks}
\label{app:modeling_metrics}

\begin{table*}[h]
    \centering
    \caption{Metrics used for the data modeling tasks in UrbanDS-Bench. An upward arrow indicates that a larger value is better, and a downward arrow indicates that a smaller value is better.}
    \label{tab:modeling_metric_mapping}
    \begin{tabular}{p{0.20\textwidth}|p{0.38\textwidth}|p{0.27\textwidth}}
        \hline
        \textbf{Task} & \textbf{Metrics} & \textbf{Evaluation values} \\
        \hline
        Land Use Classification
        & Accuracy $\uparrow$, Macro F1 $\uparrow$
        & Original class labels \\
        Crowd Flow Generation
        & MAE $\downarrow$, RMSE $\downarrow$, SMAPE $\downarrow$
        & Original outflow and inflow values \\
        OD Flow Prediction
        & CPC $\uparrow$, MAE $\downarrow$
        & Original flow values \\
        POI Check-in Prediction
        & MAE $\downarrow$, RMSE $\downarrow$, $R^2$ $\uparrow$
        & Transformed by Equation~\ref{eq:modeling_log_transform} \\
        Economic Prediction
        & MAE $\downarrow$, RMSE $\downarrow$, $R^2$ $\uparrow$
        & Transformed by Equation~\ref{eq:modeling_log_transform} \\
        Population Prediction
        & MAE $\downarrow$, RMSE $\downarrow$, $R^2$ $\uparrow$
        & Transformed by Equation~\ref{eq:modeling_log_transform} \\
        Restaurant Number Prediction
        & MAE $\downarrow$, RMSE $\downarrow$, $R^2$ $\uparrow$
        & Transformed by Equation~\ref{eq:modeling_log_transform} \\
        Takeaway Orders Prediction
        & MAE $\downarrow$, RMSE $\downarrow$, $R^2$ $\uparrow$
        & Transformed by Equation~\ref{eq:modeling_log_transform} \\
        \hline
    \end{tabular}
\end{table*}

This section describes the metrics used for the eight data modeling tasks in UrbanDS-Bench. For every task, a valid submission must contain exactly the same sample keys as the ground truth. All submitted numeric values must be finite and nonnegative. Let $N$ denote the number of evaluated values, and let $y_i$ and $\hat{y}_i$ denote the ground truth and prediction for the $i$th value, respectively. Higher values are better for Accuracy, Macro F1, CPC, and $R^2$, while lower values are better for MAE, RMSE, and SMAPE.

\subsection{Classification Metrics}

\paragraph{Accuracy.}
Accuracy measures the proportion of samples assigned the correct class:
\begin{equation}
    \mathrm{Accuracy}=\frac{1}{N}\sum_{i=1}^{N}\mathbf{1}(y_i=\hat{y}_i),
\end{equation}
where $\mathbf{1}(\cdot)$ equals 1 when its argument is true and 0 otherwise.

\paragraph{Macro F1.}
For each class $c$ in the set of ground truth classes $\mathcal{C}$, its F1 score is
\begin{equation}
    \mathrm{F1}_c=\frac{2\mathrm{TP}_c}
    {2\mathrm{TP}_c+\mathrm{FP}_c+\mathrm{FN}_c},
\end{equation}
where $\mathrm{TP}_c$, $\mathrm{FP}_c$, and $\mathrm{FN}_c$ are the numbers of true positives, false positives, and false negatives for class $c$. The score is set to 0 if the denominator is 0. Macro F1 assigns equal weight to every class:
\begin{equation}
    \mathrm{MacroF1}=\frac{1}{|\mathcal{C}|}
    \sum_{c\in\mathcal{C}}\mathrm{F1}_c.
\end{equation}

\subsection{Regression Metrics}

\paragraph{Mean Absolute Error.}
MAE is the average absolute difference between predictions and ground truth values:
\begin{equation}
    \mathrm{MAE}=\frac{1}{N}\sum_{i=1}^{N}|y_i-\hat{y}_i|.
\end{equation}

\paragraph{Root Mean Squared Error.}
RMSE gives more weight to large errors by squaring each difference before averaging:
\begin{equation}
    \mathrm{RMSE}=\sqrt{\frac{1}{N}
    \sum_{i=1}^{N}(y_i-\hat{y}_i)^2}.
\end{equation}

\paragraph{Symmetric Mean Absolute Percentage Error.}
SMAPE measures error relative to the magnitudes of both the ground truth and prediction:
\begin{equation}
    \mathrm{SMAPE}=\frac{1}{N}\sum_{i=1}^{N}
    \frac{|y_i-\hat{y}_i|}
    {( |y_i|+|\hat{y}_i| )/2+\epsilon},
\end{equation}
where $\epsilon=10^{-8}$ prevents division by zero. The implementation reports the ratio directly rather than multiplying it by 100.

\paragraph{Common Part of Commuters.}
CPC measures the overlap between two nonnegative flow distributions:
\begin{equation}
    \mathrm{CPC}=\frac{2\sum_{i=1}^{N}\min(y_i,\hat{y}_i)}
    {\sum_{i=1}^{N}y_i+\sum_{i=1}^{N}\hat{y}_i}.
\end{equation}
If both the total ground truth flow and total predicted flow are zero, CPC is defined as 1. A larger CPC indicates greater agreement between the two flow distributions.

\paragraph{Coefficient of Determination.}
The coefficient of determination compares prediction error with the variation in the ground truth:
\begin{equation}
    R^2=1-\frac{\sum_{i=1}^{N}(y_i-\hat{y}_i)^2}
    {\sum_{i=1}^{N}(y_i-\bar{y})^2},
    \qquad
    \bar{y}=\frac{1}{N}\sum_{i=1}^{N}y_i.
\end{equation}
A value closer to 1 indicates a better fit. The score can be negative when a model performs worse than predicting the ground truth mean.

\subsection{Task Specific Evaluation}

Table~\ref{tab:modeling_metric_mapping} lists the metrics used by each task. For Crowd Flow Generation, the outflow and inflow values of all region and hour pairs are flattened into one vector before MAE, RMSE, and SMAPE are calculated. For OD Flow Prediction, CPC and MAE are calculated over the flow values of all origin and destination pairs.

Five count prediction tasks use a natural logarithm before calculating MAE, RMSE, and $R^2$. Specifically, each ground truth or predicted value $v$ is transformed independently as
\begin{equation}
    g(v)=
    \begin{cases}
        \log(v), & v>1,\\
        0, & 0\leq v\leq 1.
    \end{cases}
    \label{eq:modeling_log_transform}
\end{equation}
The formulas above are then applied to $g(y_i)$ and $g(\hat{y}_i)$ rather than the original values. This transformation is used for POI Check-in Prediction, Economic Prediction, Population Prediction, Restaurant Number Prediction, and Takeaway Orders Prediction.

\end{document}